\documentclass{IOS-Book-Article}
\usepackage{algorithm} 
\usepackage{algorithmic} 
\usepackage{tikz}                                                                                                                                               
\usetikzlibrary{arrows,backgrounds,shapes}
\usepackage[T1]{fontenc}
\usepackage{mathptmx}

\usepackage{subcaption}
\usepackage{url}
\begin{document}
\begin{frontmatter}              

\title{Computational Methods for Probabilistic Inference of Sector Congestion in \\Air Traffic Management}

\author[A,C]{\fnms{Ga\'etan} \snm{Marceau Caron}%
\thanks{gaetan.marceaucaron@thalesgroup.com, gaetan.marceau-caron@inria.fr}},
\author[B]{\fnms{Pierre} \snm{Sav\'eant}
\thanks{pierre.saveant@thalesgroup.com}},
and
\author[B]{\fnms{Marc} \snm{Schoenauer}
\thanks{marc.schoenauer@inria.fr}}
\address[A]{Thales Air Systems, Rungis, France}
\address[B]{Thales Research \& Technology, Palaiseau, France}
\address[C]{TAO project-team, INRIA-Saclay Ile-de-France, Universit\'e Paris-Sud, Orsay, France}

\begin{abstract}
This article addresses the issue of computing the expected cost functions from a probabilistic model of the air traffic flow and capacity management.
The Clenshaw-Curtis quadrature is compared to Monte-Carlo algorithms defined specifically for this problem.
By tailoring the algorithms to this model, we reduce the computational burden in order to simulate real instances.
The study shows that the Monte-Carlo algorithm is more sensible to the amount of uncertainty in the system, but has the advantage to return a result with the associated accuracy on demand.
The performances for both approaches are comparable for the computation of the expected cost of delay and the expected cost of congestion.
Finally, this study shows some evidences that the simulation of the proposed probabilistic model is tractable for realistic instances.  
\end{abstract}

\begin{keyword}
Clenshaw-Curtis Quadrature; Monte-Carlo Simulation; Air Traffic Management; Probabilistic Model
\end{keyword}
\end{frontmatter}

\thispagestyle{empty}
\pagestyle{empty}

\section{Introduction}
The air traffic flow and capacity management has been extensively studied in the past years \cite{Bertsimas1998} and a review of the models used in the operational research community can be found in \cite{Agustin}. 
It covers mainly the static and the single-stage variant, which has been solved for large-scale instances of the size of the National Airspace of the United States \cite{LulliBertsimas2011}.
More recently, research effort was oriented toward anticipation of perturbations \cite{Agustn2012b} and implementing network strategies with closed-loop control \cite{DBLP:conf/cdc/NyB10} in order to mitigate the possible perturbations.
In any case, there is an interest in the research community toward the dynamic variant of the problem.
In the SESAR Joint Undertaking, the working package 07.06.05 concerns the ``dynamic Demand Capacity Balancing''.
The proposed solution for this problem is to introduce new short-term measures in the set of possible actions of the flow manager: minor ground delays, flight level capping and minor re-routings applied to a limited number of flights.
These measures ensure that the response is proportional to the magnitude of the perturbation.

One question that arises is the impact of uncertainty on the decisions.
We measured the prediction error of the tabular BADA model\cite{bada} and a steady-state flight model against real flight recordings and we observed that the amount of uncertainty changes according to the flight phases, i.e., a difference around 5-6 minutes between the climbing and the en-route phases.
This was also verified by \cite{Gilbo11} who measures a difference of 9 minutes of prediction error between proposed flights and active flights.
Besides, the Central Office for Delay Analysis Digest 2012 asserts that 36\% of the flights were delayed by 5 minutes or more and 16\% by 15 minutes or more compared to the scheduled departure time\cite{coda2012}.
The factors are numerous and the prediction errors will be a problem for the next years since some causes are still unpredictable, e.g., weather and human behavior.
For these reasons, we are interested into a non-homogeneous model of uncertainty of the system and we believe that it will enhance the robustness of the chosen optimal plan.
On the one hand, we can monitor the events in term of degree of confidence and we can take actions according to the probability of occurrence of different scenarios.
On the other hand, the probability of presence in a sector can induce some safety margins between the exit and the entrance of two different flights, i.e., reducing the time between these events will increase the probability of congestion.

\section{Model}
In this section, we briefly present the model, for more details please refer to \cite{marceau2013} and \cite{atm:sid2012a}.
First of all, it consists of two submodels, the flight model and the sector model.
The former is used to compute the expected cost of delay and the probability of presence in the sectors.
The latter takes the probability of presence for each flight and compute the expected cost of congestion.
First, the flight model is defined as follows:

\begin{equation}
  \label{eq:fpModel}
  p_N(t_N) = \int_{\Omega} \dots \underbrace{\left[ \int_{\Omega} p_{3|2}^f(t_3|t_2) \underbrace{\left[ \int_{\Omega} p_{2|1}^f(t_2|t_1) p_1^f(t_1) dt_1 \right]}_{p_2^f(t_2)} dt_2 \right]}_{p_3^f(t_3)} \dots dt_{N-1} \nonumber
\end{equation}

where $\Omega$ is the time horizon, $p_1^f(t_1)$ is the marginal probability for the flight $f$ to enter the airspace at time $t_1$ and $p_{i+1|i}^f(t_{i+1}|t_i)$ is the conditional probability to be at $i+1$ at time $t_{i+1}$ given it was at $i$ at time $t_i$. 
This equality holds because we use the Markov assumption.
Then, the probability of presence is:
\begin{equation}
\label{eq:presenceModel}
\Pr(S_{s,f}(t)) = \int_{-\infty}^t p_i^f(t) - p_{i+1}^f(t) dt
\end{equation}
which is the difference of the cumulative probabilities at the boundary points of the sector.
\newpage 
Thereafter, a stochastic process $S_{s,f}(t)$ is gathered from every flight $f$ of a sector $s$ and aggregated to obtain the stochastic process of congestion:
\begin{eqnarray}
  \label{eq:cong}
  \Pr(K_s(t) = n) &=& \sum_{|a|=n} \prod_{f \in \mathcal{F}_{|s}} \Pr(S_{s,f}(t))^{a_f} \cdot \Pr(\overline{S_{s,f}(t)})^{1-a_f} \\
  &=& \frac{1}{N_s(t)+1} \sum_{l=0}^{N_s(t)} \exp(-iwln) \dots \nonumber \\
  & & \quad \prod_{f=1}^{N_s(t)} \left[ 1-\Pr(S_{s,f}(t))+\Pr(S_{s,f}(t)) \exp(iwl) \right] \nonumber
\end{eqnarray}
where $\mathcal{F}_{|s}$ is the subset of flights crossing the sector $s$, $i=\sqrt{-1}$, $w=\frac{2 \pi}{N_s(t)+1}$, $\overline{S_{s,f}(t)}$ is the complement of $S_{s,f}(t)$ and $N_s(t) = |\{f \in \mathcal{F} | \Pr(S_{s,f}(t)) \neq 0 \}|$.
Notice that $|a|=n$ refers to the multi-index notation where $a$ is a Boolean vector and so, the sum is over all vectors with a number of 1 equals to $n$.
The Boolean $a_f$ denotes if we use the probability of the flight to be or not in the sector.
The last equality is obtained with the characteristic function of a Binomial-Poisson distribution and it can be computed with a Fast Fourier Transform \cite{hong2011}.

\begin{figure}
  \centering
  \includegraphics[scale=0.36]{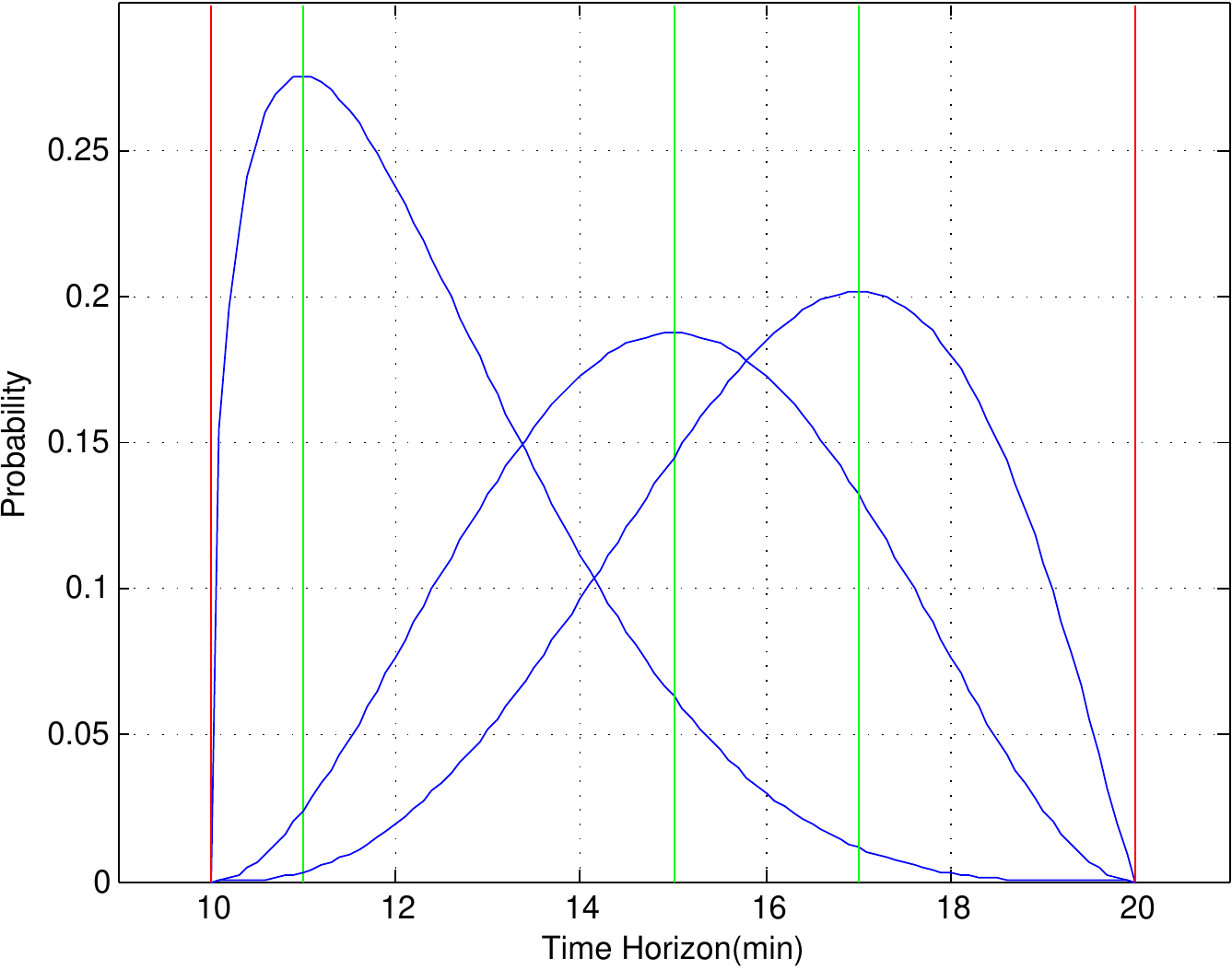}
  \caption{\label{fig:intent_model} Intent Model}
\end{figure}

The inference model needs two input models that describe the uncertainty for each flight.
The input models consist in the inbound model and the intent model.
The former is simply a marginal distribution describing the uncertainty on the arrival time in the airspace.
In this study, we consider only departures and we use the CODA digest 2012 statistics on departure delays.
The former model consists of two-dimensional functions describing the uncertainty of the arrival time at points given the time of arrival at their previous point.
Here, for the sake of simplicity, we do the assumption that this distribution is stationary but it can be removed by taking into account specific evolution of the flight state according to the boundary points, e.g., flight phases, flight intents or non-homogeneous wind field.   
To do so, expert knowledge on flight dynamics and statistical analysis from historical data seems to be the best ways to define the intent models and should be addressed in further research.
Nevertheless, in this study, we rely on the triangular and the PERT distributions, usually used to model the probability of duration of an activity in risk management. 
The PERT distribution is defined as a Beta distribution scaled on an arbitrary support:
\begin{equation}
PERT(min,max, m,\lambda) \sim min + X \cdot (max-min)
\end{equation}
where $X \sim BETA(1+\lambda \frac{m-min}{max-min}, 1+\lambda \frac{max-m}{max-min})$ and $\lambda$ is arbitrary set to 4.0, which is the value recommended in the risk management literature.
As for the triangular distribution and the Beta distribution, the support of the PERT distribution is bounded (see figure \ref{fig:intent_model}).
The bounds are obtained via the time of arrival on the previous point. 
It reflects the fact that the flight is physically limited in speed and in flying time.
Finally, we assign $m$, the mode of the distribution, to the target time of arrival.

\newpage
The probabilistic model is meant to be used in an optimization framework.
We are interested in the expected cost functions of the delay and the congestion.
These can simply be defined as an aggregation measure on every flight and sector respectively.
For the expected cost of delays, the function is:
 \begin{equation} 
\label{eq:cost1}
\mathcal{C}_1(\gamma) = \sum_{f \in \mathcal{F}} \left[ \int_{\Omega} (\tau - A_f)_+^2 \cdot p_{n^f}^f(\tau;\gamma_{|f}) d\tau \right]
\end{equation}
where  $\mathcal{F}$ is the set of flight, $A_f$ is the scheduled time of arrival and $p_{n^f}^f$ is the marginal on the last point for the flight $f$.
$\gamma_{|f}$ denotes the decision vector, the target times of arrival on each point, restricted to the targets of the flight $f$.
Indeed, in the optimization context, we are trying to find the vector $\gamma$ that minimizes the cost functions.
For the expected cost of congestion, the function is:   
\begin{equation}
\label{eq:cost2}
\mathcal{C}_2(\gamma) = \int_{\bar{\Omega}} \sum_{s \in \mathcal{S}} \left[ \sum_{n=C_s(\tau)+1}^{N_s(\tau)} (n - C_s(\tau))^{2} \cdot \Pr(K_s(\tau)=n;\gamma) d\tau \right] \nonumber
\end{equation}
where $\mathcal{S}$ is the set of sectors, $C_s(\tau)$ is the capacity of the sector $s$ at time $\tau$.
In the following, this will be set to a constant function.
Here, we use quadratic cost functions to avoid the problem of fairness, where a flight or a sector is continuously penalized for the benefit of the others. 

\section{Algorithms}
In this section, the Clenshaw-Curtis quadrature is compared with the Monte-Carlo approach for the simulation of the probabilistic model and are tested on two benchmarks, one per cost function, in order to emphasize the differences between computational time and accuracy.

The Clenshaw-Curtis quadrature is an adaptive numerical integration technique. 
The choice of the algorithm was mainly oriented toward generic algorithms, since there are no assumptions on the density functions.
However, the marginalization process consists in an integral of a product of functions and so, the Clenshaw-Curtis Quadrature is a good candidate.
It is based on the expansion of the integrand in terms of Chebyshev polynomials and it is recognized for its efficiency and its accuracy.
In our case, for one flight, the number of dimensions is equal to the number of sectors plus one, typically a value around ten.
Consequently, integrating the cost function with the last marginal directly is impractical due to the {\it curse of dimensionality}.
A straightforward solution is to stock the intermediate results at each node, from the first marginal to the last (cf. equation \ref{eq:fpModel}).
Then, the cost function is integrated with the approximation of the last marginal.
The resulting error depends directly on the discretization required to stock the intermediate functions.
A finer discretization will lead to more evaluation points at each node and consequently, the computational burden will increase exponentially (see figure \ref{fig:ni_time}).
A coarse discretization will generate a larger error, which will be propagated at each node, and so, accuracy will be easily lost.
The problem of dimensionality concerns mainly the evaluation of the expected cost of delay for the flights.
The evaluation of the expected cost of congestion is different since it requires intermediate operations.
The first one is the computation of the probability for a flight to be in the sector at a given time (see equation \ref{eq:presenceModel}).
The second one gathers the results from the previous one for each flight, defines a Binomial-Poisson distribution and computes its probability mass function (see equation \ref{eq:cong}).
The last one is the integration of the stochastic process defined by the result of the previous step with the cost function (see equation \ref{eq:cost2}).

The Monte-Carlo approach is fundamentally different from numerical integration.
It consists in simulating the system many times to obtain a statistic on the expected value of the distribution.
By the law of large numbers, the sample mean converges in probability and almost surely to the expected value.
Moreover, by the Central Limit Theorem, the distribution associated to $\frac{\sqrt{n}(\overline{x} - \mu)}{\sigma}$ converges in distribution to a Normal Distribution where $\overline{x}$ is the sample mean, $\mu$ is the expected value and $\sigma^2$ is the variance of the underlying distribution.
As a consequence of this important result, the estimate of the mean will be better if the variance of the underlying distribution decreases or if the number of samples increases.
Nevertheless, as $n$ becomes large, the impact of the latter on the reduction of the error of the mean will decrease because of the square root function at the numerator.

Our system is described by a huge joint distribution.
Nevertheless, solely a few marginals are of interest; those which captures the important characteristics of the airspace.
On the one hand, the expected cost of delay needs to be computed for each flight.
Since we assume that they are mutually independent given the objectives, samples can be drawn independently from the marginal distribution of the last point for every flight.
Because of the Markov assumption, this can be done easily with the forward sampling algorithm. 
On the other hand, the expected cost of congestion needs to be computed for each sector.
The sectors are also mutually independent given the objectives, but the congestion depends on the flights.
However, since the distributions of congestion are not identical from one sector to another, the number of samples required to attain a given accuracy may not be equal. 
It seems a good idea to determine empirically the number of samples per sector in order to give more importance to the ones with high variance.
The standard error of the mean is the simplest estimate of the true standard deviation of the mean.
From this measure, let $\overline{\epsilon_r}$ be a threshold on the estimate of the relative error, computed by the ratio of the standard error and the sample mean, and $\epsilon_a$ be a threshold on the absolute error.

\begin{algorithm}
\caption{\label{alg:ccs} Congestion Cost Sampling}
\begin{algorithmic}[1]
  \REQUIRE $s, p$, $\mathcal{F}_{|s}, C_s$
  \FOR{$f \in \mathcal{F}_{|s}$} 
  \STATE $\xi_i = ForwardSampling(f,p,s)$
  \ENDFOR
  \STATE $T := \emptyset$
  \FOR{$i=0 \to |\mathcal{F}_{|s}|$} 
  \STATE $T = T \cup (\underline{\xi_i},1) \cup (\overline{\xi_i},-1)$
  \ENDFOR
  \STATE SORT(T,time,ascending)
  \STATE $a:=1$
  \STATE $\tau := T(0)_0$
  \STATE $M := \emptyset$
  \FOR{$t \in T \backslash \{T(0)\}$}
  \STATE $M[a] = M[a] \cup (\tau,t_0)$
  \STATE $\tau = t_0$
  \STATE $a = a + t_1$
  \ENDFOR
  \STATE $J_{cong} := 0$
  \STATE $\overline{a} = \max_i\{M[i] \neq \emptyset\}$ 
  \FOR{$i= C_s+1 \to \overline{a}$}
  \IF{$M[i] \neq \emptyset$}
  \FOR{$t \in M[i]$}
  \STATE $J_{cong} = J_{cong} + cost(i) \cdot (\overline{t} - \underline{t})$
  \ENDFOR
  \ENDIF
  \ENDFOR
  \RETURN $J_{cong}$
\end{algorithmic}
\end{algorithm}
In this study, {\it particle} refers to the times of overfly, generated by the Monte-Carlo routine, necessary for the computation of the congestion of a sector.
Algorithm \ref{alg:ccs} is a sampling routine giving the cost of congestion for one particle identified by $p$.
This identifier is used to recover the scenario from the flight model, through the ForwardSampling routine, and if the particle already exits, it can be retrieved from the memory and if not, it can be generated by the forward sampling algorithm and stored.
This permits to create partial scenario according to the convergence of each sector and hence, to reduce the computational burden.
Also, even if the forward sampling picks a value for every point of the flight plan, the routine returns only the two entry/exit points associated to the sector $s$.
Consequently, $\xi_i = [\underline{\xi_i}, \overline{\xi_i}]$ is the time interval when the flight is in the sector.
Also, we use an ordered set $T$ of pairs to stock the timestamp and the event of entrance (+1) and of exit (-1).
$T(i)_j$ denotes the ith pair of the set and the jth element of this pair.
Line 9 requires that we sort the set $T$ according to the time in an ascending order.
This corresponds to the idea of a sweep line algorithm.
Since the size of $T$ is twice the number of aircraft in the sector, the algorithm needs to sort only a few items.
Then, by counting the number of entrance/exit at each event, we obtain a time interval with a corresponding number of flights.
Hereafter, with a multimap $M$, a map with multiple values for one key, the intervals are stored according to the number of flights.
When the sweep algorithm terminates, the cost is determined by summing the cost function of the number of flights (the keys of $M$) multiplied by the length of the time interval (the values of $M$).
The complexity of the routine depends solely on the number of flights and by avoiding the discretization of the time horizon, the computational error is bounded by the machine precision.

Thereafter, this routine is called several times by the following Monte-Carlo routine, which ensures the convergence of the process toward the mean value with a soft condition according to the relative and absolute standard error of the mean. 
\begin{algorithm}
  \caption{\label{alg:eccmc} Expected Cost of Congestion with Monte-Carlo}
  \begin{algorithmic}[1]
    \REQUIRE $\mathcal{F}_{|s}, C_s, \epsilon_{rel}, \epsilon_{abs}, n_{init}$
    \FOR{$i = 0 \to n_{init}$}
    \STATE $J_{cong}^{(i)} = CongestionCostSampling(i,\mathcal{F}_{|s},C_s)$
    \STATE $acc_{\overline{x}}(J_{cong}^{(i)})$
    \STATE $acc_{SD_{\overline{x}}}(J_{cong}^{(i)})$
    \ENDFOR
    \STATE $ n := n_{init} + 1$
    \WHILE{$SD_{\overline{x}} > \epsilon_{rel} \cdot \overline{x} \quad \wedge \quad  SD_{\overline{x}} < \epsilon_{abs}$}
    \STATE $J_{cong}^{(n)} = CongestionCostSampling(n,\mathcal{F}_{|s},C_s)$
    \STATE $\overline{x} = acc_{\overline{x}}(J_{cong}^{(n)})$
    \STATE $SD_{\overline{x}} = acc_{SD_{\overline{x}}}(J_{cong}^{(n)})$
    \STATE $n = n + 1$
    \ENDWHILE
  \end{algorithmic}
\end{algorithm}
Algorithm \ref{alg:eccmc} uses the concept of accumulator(acc) which computes the mean $\overline{x}$  and the standard error of the mean $SD_{\overline{x}}$ in an online way.

\begin{algorithm}
  \caption{\label{alg:cmmc}Congestion Monitoring with Monte-Carlo}
  \begin{algorithmic}[1]
    \REQUIRE $s, F$
    \STATE $\mathcal{I} := \emptyset$
    \COMMENT{A set for accumulating the congested intervals}
    \FOR{$i = 0 \to n_{init}$}
    \STATE ...
    \COMMENT{Initialization of the accumulators: refer to line 7 to 40}
    \ENDFOR
    \STATE $n := n_{init} + 1$
    \WHILE{$max(E) > \epsilon_{rel}$}
    \STATE $E = \emptyset$
    \STATE $I = CongestionSampling(s,n,F_{|s})$
    \FOR{$t \in I$}
    \IF{$M[\underline{t}] = \emptyset$}
    \STATE $M[\underline{t}] = acc_{\overline{x}}$
    \FOR{$\xi \in \mathcal{I}$}
    \IF{$\underline{t} \in \xi$}
    \STATE $M[\underline{t}](1)$
    \ELSE
    \STATE $M[\underline{t}](0)$
    \ENDIF
    \ENDFOR
    \ELSIF{$M[\overline{t}] = \emptyset$}
    \STATE $M[\overline{t}] = acc_{\overline{x}}$
    \FOR{$\xi \in \mathcal{I}$}
    \IF{$\overline{t} \in \xi$}
    \STATE $M[\overline{t}](1)$
    \ELSE
    \STATE $M[\overline{t}](0)$
    \ENDIF
    \ENDFOR
    \ENDIF
    \STATE $l = lower(M,\underline{t})$
    \STATE $u = upper(M,\overline{t})$ 
    \FOR{$\tau \in key(M)$}
    \IF{$\tau < l \vee \tau => u$}
    \STATE $E = E \cup M[\tau](0)_{\overline{x}}$
    \ELSE
    \STATE $E = E \cup M[\tau](1)_{\overline{x}}$
    \ENDIF
    \ENDFOR
    \ENDFOR
    \STATE $\mathcal{I} = \mathcal{I} \cup I$
    \STATE $n = n + 1$
    \ENDWHILE
    \RETURN M
  \end{algorithmic}
\end{algorithm}
Finally, algorithm \ref{alg:cmmc} returns a stochastic process where each point contained in the mapping $M$ has converged in terms of error of the mean.
As a matter of fact, one accumulator is created per point (see line 10,19) and updated with the previous samples stocked in $\mathcal{I}$.
The sub-routine $CongestionSampling$ gives a set of congested temporal intervals associated to the particle $n$.
It can be implemented similarly to the algorithm \ref{alg:ccs}, but we stock only the temporal intervals associated to a number of flights greater than the capacity.
For this reason, the sub-routine $CongestionSampling$ returns a finite set of temporal intervals $I$.
Then, the algorithm tests if the boundary points of the congested interval $t = [\underline{t}, \overline{t}]$ are in the map (line 9,18).
Because these timestamps are real values, we need to define a coefficient $\epsilon$, which will manage the trade-off between the number of points in the map and the accuracy.
So, a timestamp $x$ is considered to be in the map if there exists a timestamp $y$ already in the map where $|x-y| < \epsilon$.  
We consider this algorithm to be adaptive in the sense that it will choose a number of points according to $\epsilon$ and to the complexity of the underlying stochastic process.
To illustrate this idea, a purely deterministic system requires $2n$ points where $n$ is the number of congested temporal intervals.
Indeed, we need the transition moments between probability 0 and 1 and the Monte-Carlo routine will always return these points.
Let $t_i^*$ be the timestamps of the boundary of the congested intervals.
When uncertainty is considered, the Monte-Carlo routine return points in the range $t_i^* \pm \Delta t_i$ where $\Delta t_i$ is the supremum of the interval length when the probability of presence fluctuates for each flight contributing to the congestion.
Consequently, the probability of congestion fluctuates too and this is where the interpolation points are created in order to capture the non-linearity of the evolution.
Outside of these ranges, the probability is either 0 or 1, and so the interpolation is trivial.
Besides, the algorithm requires that the data structure $M$ implements the functions lower(t) and upper(t), which return the first element whose key is not considered to be before and the first element whose key is considered to be after $t$ respectively.
These operations are generally provided with binary search tree.

\section{Results}
All the results were computed on an Intel\textsuperscript{\textregistered} Core\textsuperscript{\texttrademark} i7-3770K CPU with 8 x 3.50GHz and 15,6 Go of memory.
The programs are implemented in C++11 with the help of the statistical library from Boost 1.53, the Clenshaw-Curtis quadrature implemented in the {\it Gnu Scientific Library 1.15} and FFTW \cite{FFTW05}. 

\begin{figure}
  \centering
  \begin{subfigure}[b]{0.5\textwidth}
    \centering
    \includegraphics[scale=0.36]{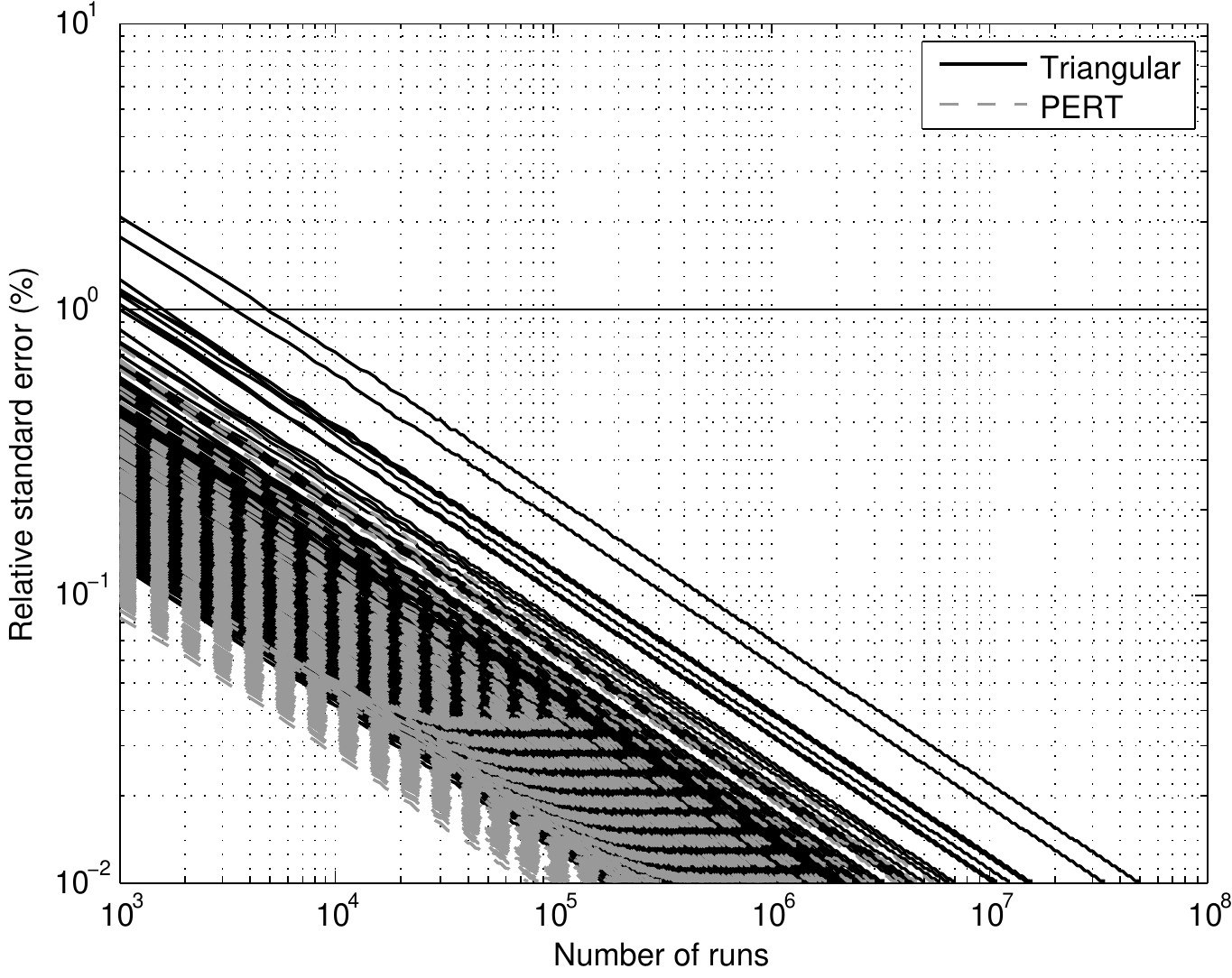}
    \caption{\label{fig:mc_convergence}Convergence}
  \end{subfigure} 
  \qquad
  \begin{subfigure}[b]{0.4\textwidth}
    \centering
    \includegraphics[scale=0.38]{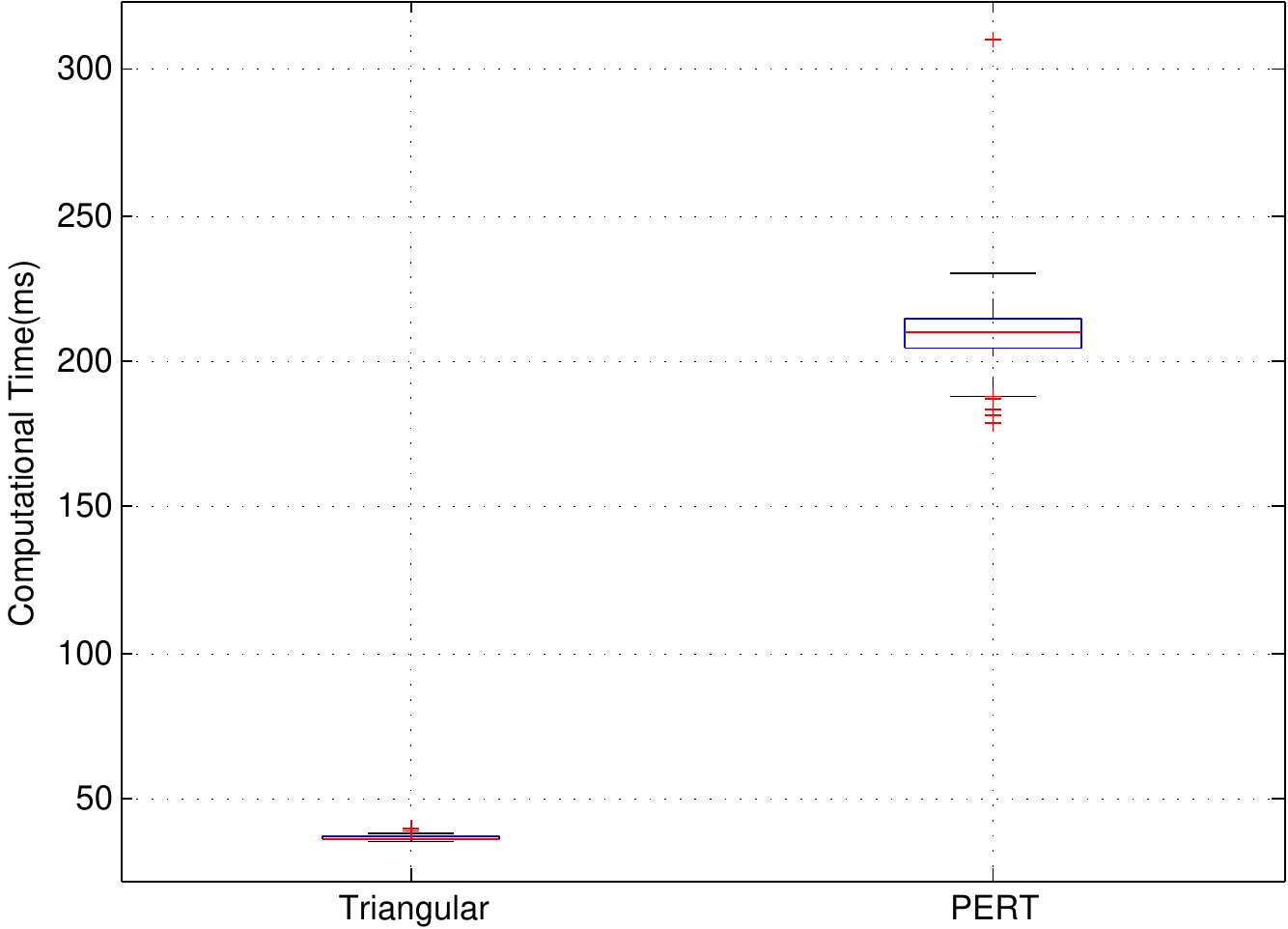}
    \caption{\label{fig:mc_time} Computational time}
  \end{subfigure}
  \caption{\label{fig:mc} Monte-Carlo Performances for the Expected Cost of Delays}
\end{figure}

We will assess of the differences between the two approaches with a flight crossing 11 sectors, which is realistic for the scope of our airspace.
Simulations are conducted with both the triangular and the PERT distributions.
The support for the initial distribution is fixed to the delays on departure given in the CODA digest 2012, fitted with a continuous linear function,  and the supports for conditional distributions are fixed to 180 seconds.
Because the supports of the two distributions are the same, the PERT distribution possesses a lower variance than the triangular distribution. 
The flight can takeoff at any time between 5 minutes and 60 minutes, every sector requires ten minutes to be crossed and the target time of arrival is around 130 minutes.
Then, we compute the expected cost of delay for 300 decision vectors randomly drawn from an uniform distribution covering the feasible set.
To obtain a good estimate of this value, we use Monte-Carlo simulations with a relative error of $10^{-4}$ and an absolute error of $0.1$.
Figure \ref{fig:mc_convergence} shows the Monte-Carlo evolution for each run.
First, the evolution for every run is linear in the log-log scale.
The reason is that the estimation of the standard deviation does not fluctuate with more samples.
So, the decrease in the relative standard error is only driven by the square root of the number of samples.
The second remark concerns the difference between the triangular and the PERT distribution.
Since the overlapping between the runs is important, it is not clear that one distribution requires more computations than the other.
\begin{figure}
  \centering
  \begin{subfigure}[b]{0.5\textwidth}
    \centering
    \includegraphics[scale=0.36]{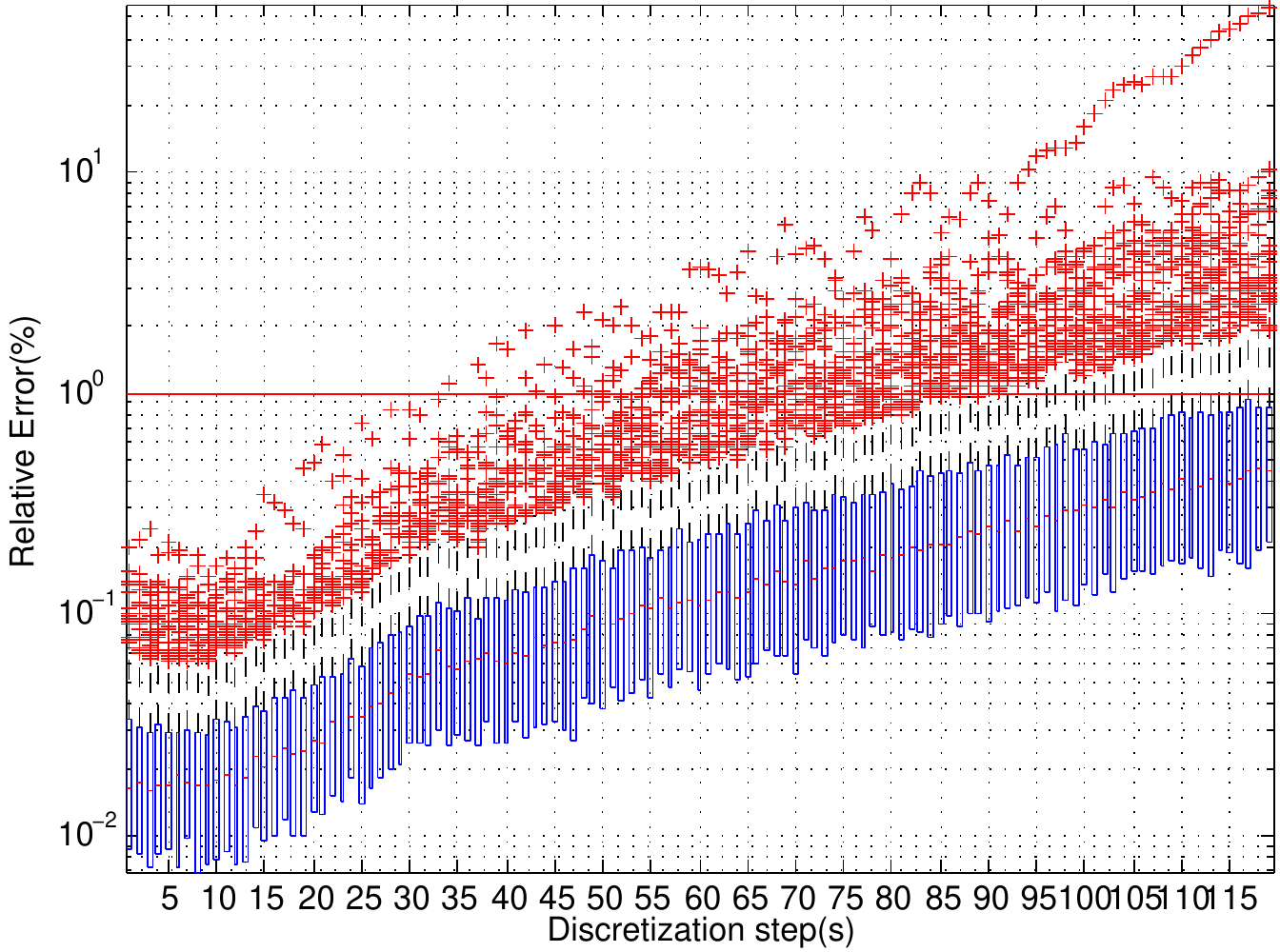}
    \caption{\label{fig:ni_accuracy_trian} Triangular Distribution}
  \end{subfigure} 
  \qquad
  \begin{subfigure}[b]{0.4\textwidth}
    \centering
    \includegraphics[scale=0.38]{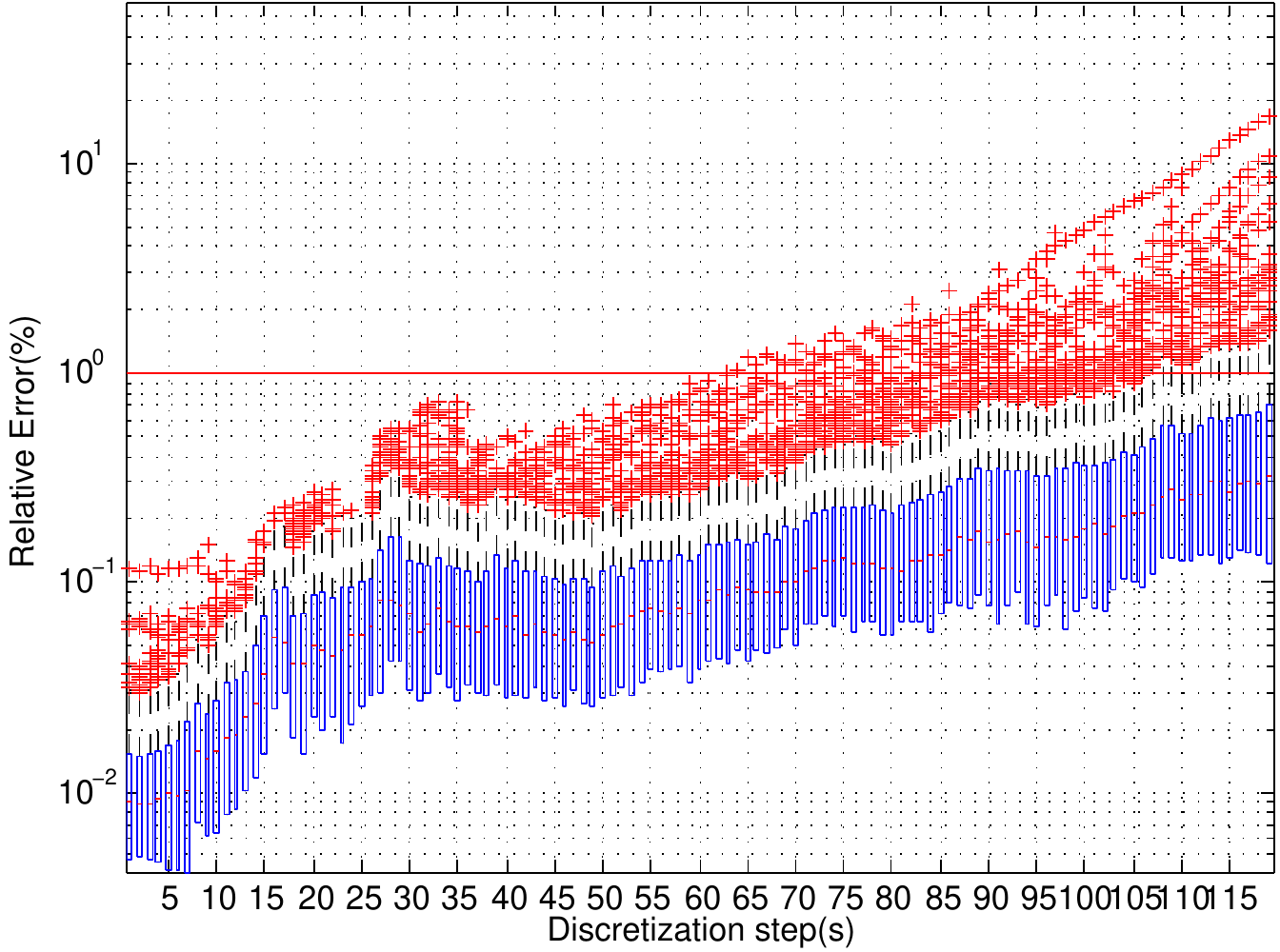}
    \caption{\label{fig:ni_accuracy_pert} PERT Distribution}
  \end{subfigure}
  \caption{\label{fig:ni_accuracy} Clenshaw-Curtis Quadrature Accuracy for the Expected Cost of Delays}
\end{figure}
\begin{figure}
  \centering
  \includegraphics[scale=0.36]{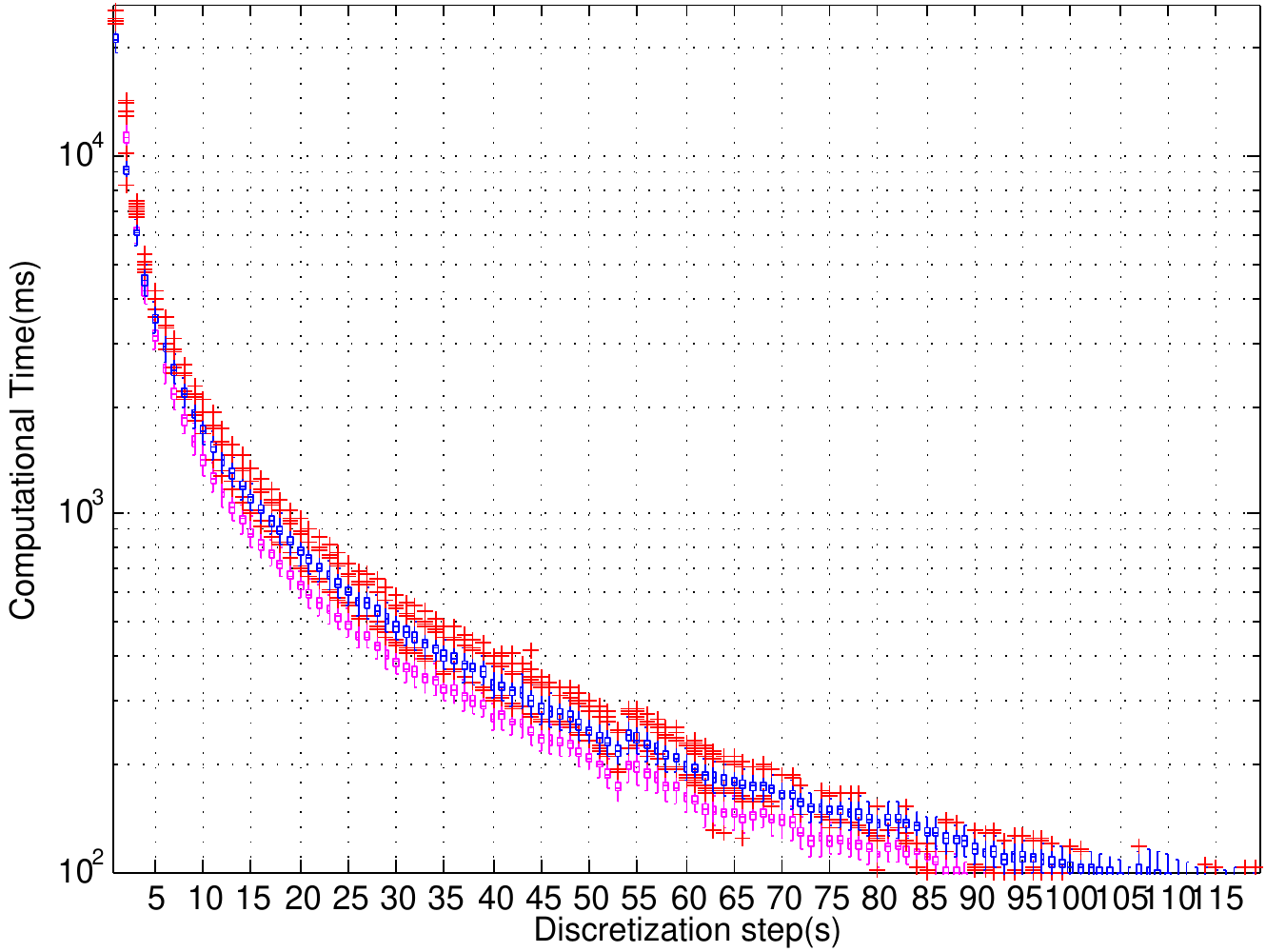}
  \caption{\label{fig:ni_time}Computational Time for Clenshaw-Curtis Quadrature}
\end{figure}
\begin{figure}
  \centering
  \begin{subfigure}[b]{0.5\textwidth}
    \centering
    \includegraphics[scale=0.36]{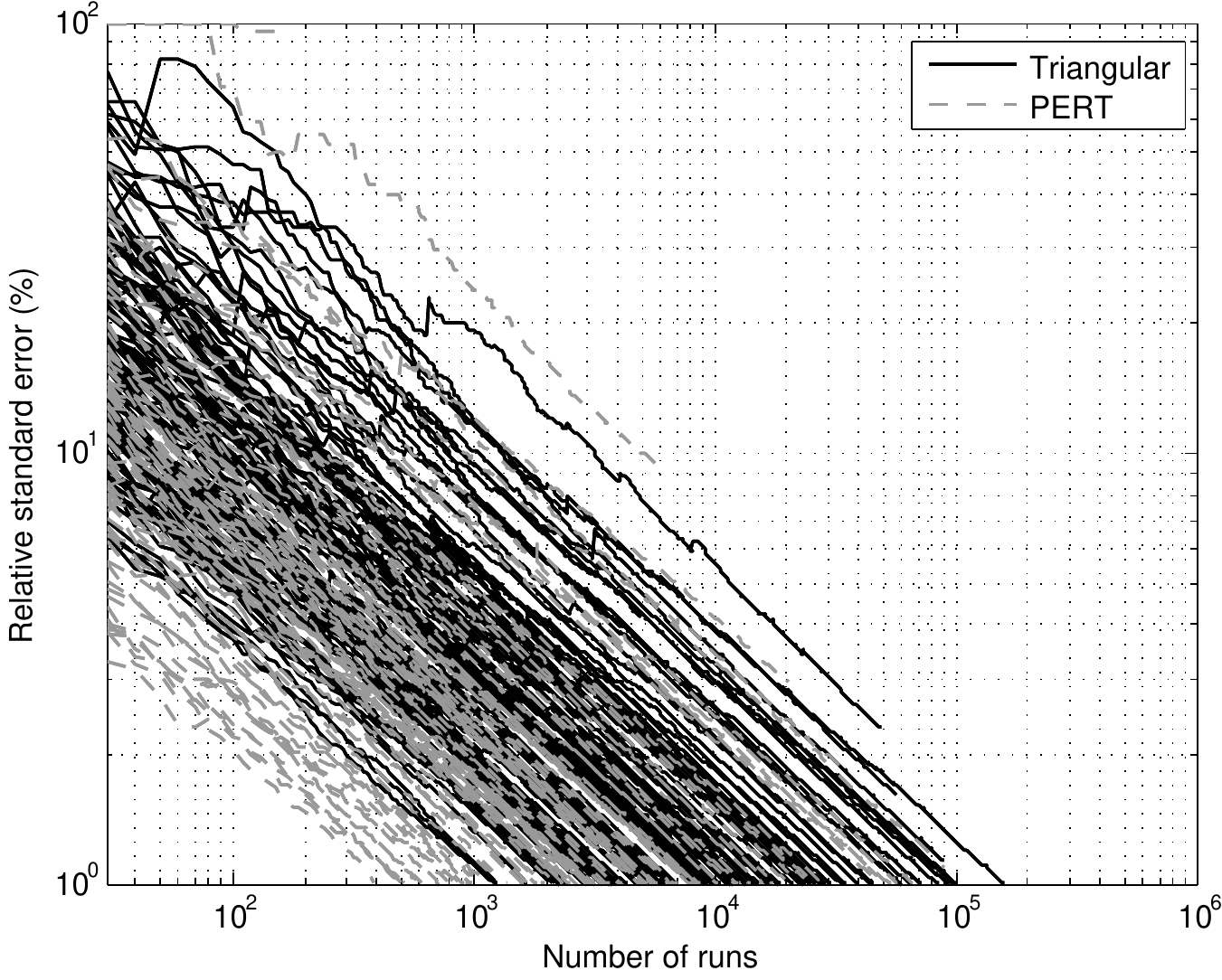}
    \caption{\label{fig:mc_convergence_cong}Convergence}
  \end{subfigure} 
  \qquad
  \begin{subfigure}[b]{0.4\textwidth}
    \centering
    \includegraphics[scale=0.38]{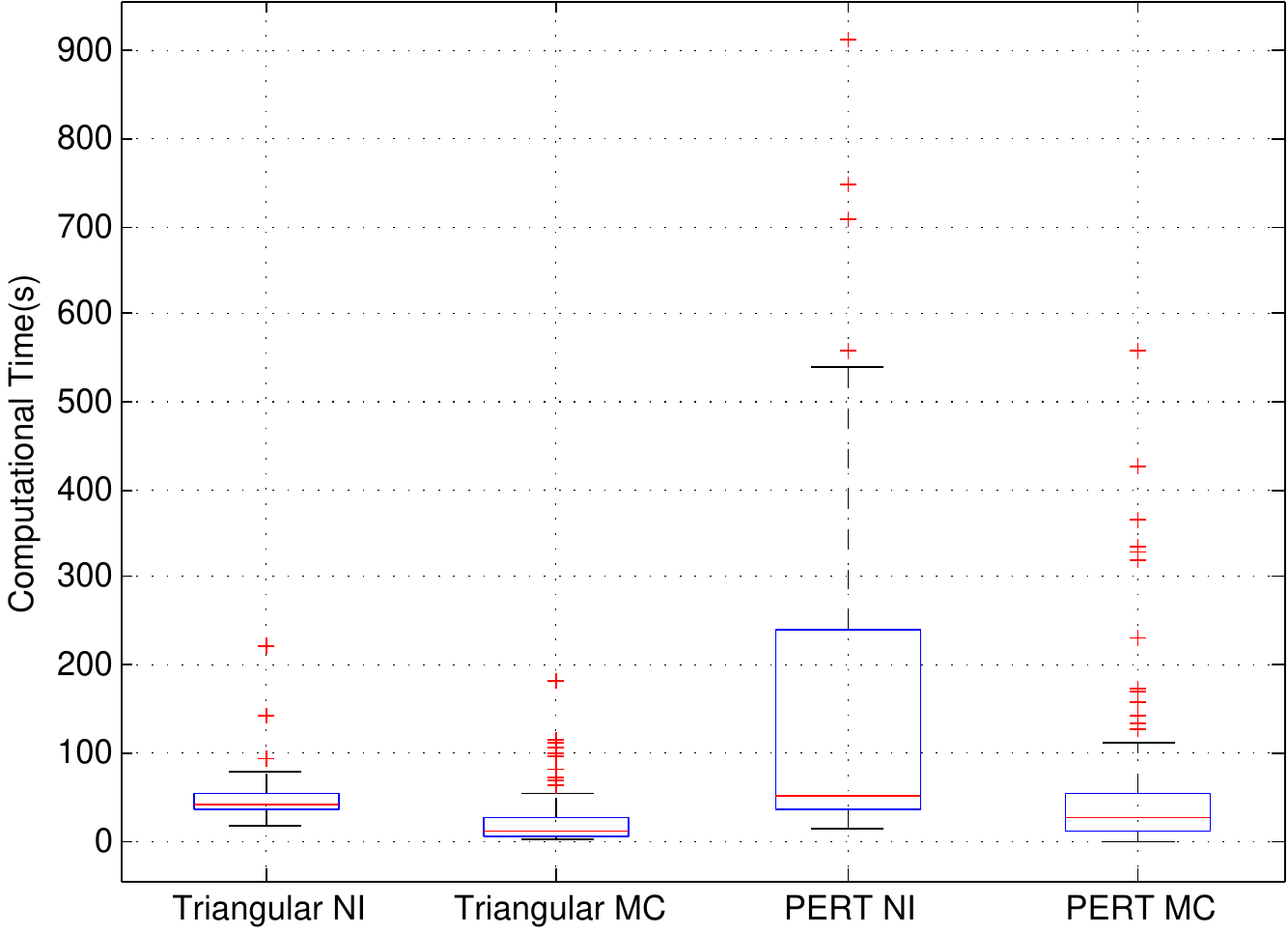}
    \caption{\label{fig:mc_ni_time} Computational time}
  \end{subfigure}
  \caption{\label{fig:mc_ni_cong} Monte-Carlo and Clenshaw-Curtis Quadrature Performances for the Expected Cost of Congestion}
\end{figure}
Nevertheless, the computational time required to generate one sample for the PERT distribution is around eight times higher than the triangular distribution, i.e., 309,19 microseconds against 38,47 microseconds.
Figure \ref{fig:mc_time} shows the computational time required to obtain a relative error of 1\%.
Consequently, the use of the PERT distribution can only be justified by a gain in modeling precision. 

For the Clenshaw-Curtis quadrature, the results were compared to the Monte-Carlo routine with a relative error of 0.01\%.
First, the two approaches converge to the same value, with a maximum error of 0.3\% for a discretization step of 1 second (cf. figure \ref{fig:ni_accuracy}).
Then, as the discretization step becomes larger, the error increases slowly and reaches the accuracy threshold of 1\% around 30s and 65 seconds for each distribution respectively.
For the computational time, the numerical integration method is more stable to the change of distribution (cf. figure \ref{fig:ni_time} and figure \ref{fig:mc_time}).
For the PERT distribution, the computational times is hardly distinguishable between the two approaches.
Moreover, we have examined that the Monte-Carlo performances are greatly dependent on the amount of uncertainty where this is not the case for the numerical integration.
Therefore, we believe that over a threshold of uncertainty, the numerical method can surpass the forward sampling in computational time. 
Nevertheless, it was also found that the forward sampling always overestimates the error of the mean and so, there are still opportunities for further enhancements of the estimator with more sophisticated Monte-Carlo routines.

In the same manner, we compare the two approaches for the computation of the expected cost function of congestion.
We verify that the two methods converge to the same result with a 1\% accuracy error for 100 random vectors drawn uniformly over the feasible set.
Here, we can see that the estimation of the standard error of the mean depends on the number of runs (see figure \ref{fig:mc_convergence_cong}), contrary to the previous estimation.
Also, there is some runs that converge to the absolute error, fixed at 0.1.
It was found that these runs compute a low congestion cost and need a lot of samples to gain little accuracy.
The same difficulty appears for the computational time of the numerical integration with PERT distribution (see figure \ref{fig:mc_ni_time}).
Since the PERT distribution has a lower variance, the expected cost function of congestion is lower than the triangular distribution and for certain decision vectors, the method has difficulties to converge to the required accuracy.
For this reason, we think that the use of the absolute error threshold is recommended to avoid spending time in the computation of low congestion cost.

\begin{figure}
  \centering
  \includegraphics[scale=0.36]{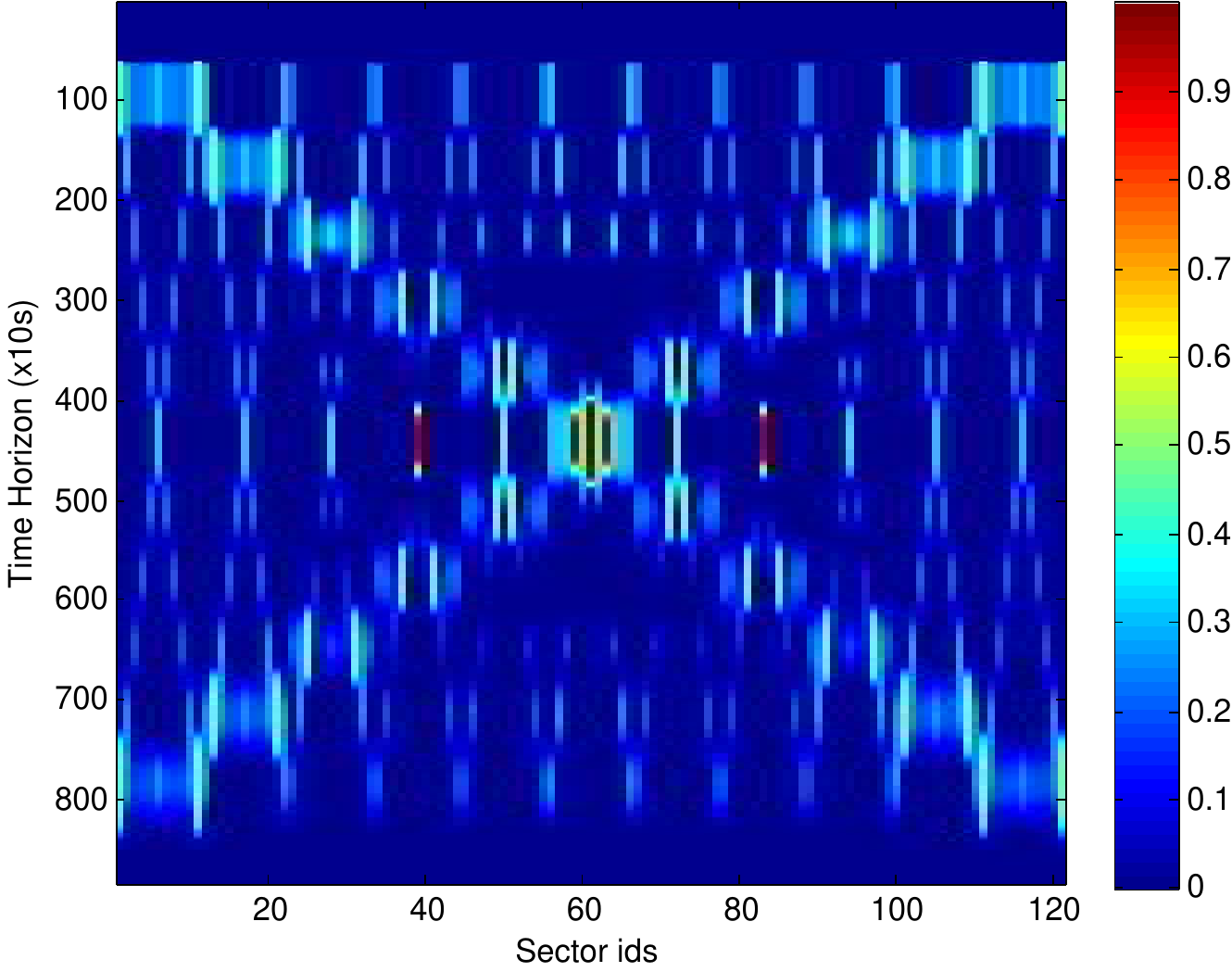}
  \caption{\label{fig:visu_cong} Congestion Visualization along Time}
\end{figure}

Finally, algorithm \ref{alg:cmmc} was used to generate figure \ref{fig:visu_cong} where each column give the evolution of the congestion probability for a sector (from top to down) on a time horizon of 2 hours and a half.
The instance consists of 121 sectors disposed as a 11x11 grid where 192 flights arrive from every direction and are going toward the opposite direction.
The capacities were arbitrary chosen to induce a congestion probability greater than zero and were assigned to the sectors in order to obtain the symmetry observed in the figure.
The result was generated in 219 seconds and so, we believe that computing the probability of congestion with an accuracy around 1\% is tractable for large instances.

\newpage
\section{Conclusion}
This article presents a comparative between the Clenshaw-Curtis method and Monte-Carlo routines using a sweep-line algorithm.
We were interested in the computation of two cost functions concerning the expected delays and the expected congestion.
Also, a Monte-Carlo routine for congestion monitoring is defined.
For the computation of the cost function, no discretization is used and so, the result can be approximate arbitrary well depending on the chosen number of particles.
For the congestion monitoring, the algorithm determines the points where the probability of congestion fluctuates.
According to the result, the Monte-Carlo performances are mitigated as the uncertainty increases where it seems not be the case for the numerical integration.
Consequently, we recommend to perform a comparative with random vectors when changing the parameter of the model before the optimization in order to choose the right approach.
However, one advantage of the Monte-Carlo simulations is the ability to perform partial simulation, i.e., to return a result and the associated accuracy in a limited amount of time.
Also, the Monte-Carlo routine for monitoring the congestion was able to complete in a reasonable amount of time for a large instance.
In summary, both approaches can be used to evaluate the cost functions on this probabilistic model, which will be required in the following studies concerning optimization.     

\bibliographystyle{plain}
\bibliography{library}

\end{document}